\pdfoutput=1

\documentclass[11pt]{article}

\usepackage[final]{acl}
\usepackage{rotating}  

\usepackage{times}
\usepackage{latexsym}

\usepackage[T1]{fontenc}
\usepackage{booktabs}

\usepackage[utf8]{inputenc}

\usepackage{microtype}

\usepackage{inconsolata}

\usepackage{graphicx}
\usepackage{pdflscape}
\usepackage{array}

\usepackage[frozencache, cachedir=minted-cache]{minted}

\setminted{
    frame=single,
    fontsize=\footnotesize,
    breaklines
}

\newcolumntype{L}[1]{>{\raggedright\let\newline\\\arraybackslash\hspace{0pt}}m{#1}}
\newcolumntype{C}[1]{>{\centering\let\newline\\\arraybackslash\hspace{0pt}}m{#1}}
\newcolumntype{R}[1]{>{\raggedleft\let\newline\\\arraybackslash\hspace{0pt}}m{#1}}

\title{NeuralNexus at BEA 2025 Shared Task: Retrieval-Augmented Prompting \\ for Mistake Identification in AI Tutors}

\author{\textbf{Numaan Naeem}\quad  
        \textbf{Sarfraz Ahmad} \quad
        \textbf{Momina Ahsan} \quad
        \textbf{Hasan Iqbal} \quad \\
MBZUAI \quad \\
\texttt{\{numaan.naeem, sarfraz.ahmad, momina.ahsan, hasan.iqbal\}@mbzuai.ac.ae}
}

\begin{document}
\maketitle

\begin{abstract}
This paper presents our system for \textsc{\textbf{Track 1: Mistake Identification}} in the \textsc{BEA 2025 Shared Task on Pedagogical Ability Assessment of AI-powered Tutors}. The task involves evaluating whether a tutor’s response correctly identifies a mistake in a student’s mathematical reasoning. We explore four approaches: (1) an ensemble of machine learning models over pooled token embeddings from multiple pretrained langauge models (LMs); (2) a frozen sentence-transformer using \textsc{[CLS]} embeddings with an MLP classifier; (3) a history-aware model with multi-head attention between token-level history and response embeddings; and (4) a retrieval-augmented few-shot prompting system with a large language model (LLM) i.e. \textsc{GPT 4o}. Our final system retrieves semantically similar examples, constructs structured prompts, and uses schema-guided output parsing to produce interpretable predictions. It outperforms all baselines, demonstrating the effectiveness of combining example-driven prompting with LLM reasoning for pedagogical feedback assessment. Our code is available at \url{https://github.com/NaumanNaeem/BEA_2025}.
\end{abstract}

\section{Introduction}

Conversational AI systems are increasingly being used for educational applications, particularly in the form of AI-powered tutors that can engage students in instructional dialogues. While recent advances in LLMs have made it possible to generate fluent and context-aware responses, evaluating whether these responses exhibit true pedagogical ability remains a fundamental challenge. Traditional dialogue evaluation metrics, such as fluency, coherence, or BLEU-like scores, fall short in capturing educational effectiveness, such as whether the tutor correctly identifies a student’s mistake or provides helpful, targeted feedback.

The \textsc{BEA 2025 Shared Task on Pedagogical Ability Assessment of AI-powered Tutors} \cite{kochmar-etal-2025-bea} addresses this gap by introducing a standardized evaluation benchmark and taxonomy to assess pedagogical abilities in AI-generated tutor responses. In particular, \textsc{Track 1: Mistake Identification} focuses on determining whether a tutor's response correctly detects and communicates an error in the student’s reasoning within a mathematical dialogue. The benchmark used in this task is based on \textsc{MRBench} \cite{maurya-etal-2025-unifying}, which includes \textit{192} dialogues and over \textit{1,500} responses from human and LLM tutors, annotated across \textit{eight} pedagogical dimensions grounded in learning sciences.

\section{Methodology}

We tackle \textsc{Track 1: Mistake Identification}, which involves determining whether a tutor's response correctly identifies a student's mistake in a multi-turn mathematical dialogue. Given the subtle and varied nature of student errors and tutor feedback, this task demands both contextual understanding and pedagogical sensitivity. To address this, we developed and evaluated \textit{four} distinct approaches: three baseline models leveraging traditional classification techniques and transformer embeddings, followed by a final retrieval-augmented few-shot classification technique using LLMs.

\subsection{Layered Embedding Extraction with Classical ML Ensemble}
In our first baseline, we designed a layered ensemble approach by extracting embeddings from several pre-trained transformer models, including \textsc{BERT}, \textsc{RoBERTa}, \textsc{XLNet}, \textsc{T5}, and \textsc{GPT-2}. To handle this flexibly, we developed a unified \textsc{LM\_EMBED} class that tokenizes and encodes both conversation history and tutor responses using each model's specific configuration. We applied average pooling over the token embeddings to produce fixed-length vectors for each input, and then averaged the conversation and response vectors to create the final input representation. Using these features, we trained a diverse set of traditional classifiers i.e. SVM, Decision Tree, Random Forest, Logistic Regression, Naive Bayes, KNN, AdaBoost, and MLP, each optimized using \textsc{GridSearchCV} with 10-fold cross-validation. We then built a meta-classifier by stacking the prediction probabilities from these base models and training a logistic regression model on top. This ensemble strategy allowed us to combine the strengths of different embedding models and classifiers, leading to more stable and accurate predictions compared to using any single model alone.

\subsection{Token-Level Attention with History-Aware Model}
In our second baseline, we modeled the interaction between the conversation history and tutor response using a token-level attention mechanism without any pooling during embedding extraction. We used a transformer encoder (\texttt{sentence-transformers/all-mpnet-base-v2}) to obtain full token-level representations for both the conversation history and the tutor's response. These representations were then passed into a custom multi-head attention module. Specifically, we treated the response as the query (Q) and the history as both the key (K) and value (V) in a standard multi-head attention setup. The output of the attention layer was mean-pooled along the sequence length dimension, and a small feedforward network mapped the pooled vector to three output classes. The model was trained using cross-entropy loss with the \textsc{AdamW} optimizer, and predictions were generated by taking the argmax over the logits. This architecture allows the model to explicitly attend to relevant parts of the history when interpreting the tutor’s response, resulting in a more nuanced classification of pedagogical mistakes.

\subsection{Frozen Sentence-Transformer with MLP Classifier}
Our third baseline models the pedagogical mistake identification task as a supervised classification problem using fixed sentence embeddings. We use a frozen sentence-level transformer model (\texttt{sentence-transformers/all-mpnet-base-v2}) to independently encode the conversation history and the tutor's response, extracting the \texttt{[CLS]} token from the final hidden state as a dense representation. These embeddings are projected through two separate linear layers and concatenated to form a joint feature vector, which is passed to a shallow feedforward neural network to predict one of three mistake identification categories. We trained this model using cross-entropy loss and the \textsc{AdamW} optimizer, keeping the encoder frozen throughout training. To improve efficiency, we cached the embeddings as \texttt{.npz} files. The final output was restructured to match the original JSON format for evaluation, preserving conversation IDs, model names, tutor responses, and predicted mistake annotations.

\subsection{Retrieval-Augmented Few-Shot Classification with LLM-as-a-Judge}
Our final and most effective approach tackles the mistake identification task as a judgment problem, using a retrieval-augmented few-shot prompting strategy powered by large language models (LLMs). Instead of training a traditional classifier, we designed a modular pipeline built with LangChain. At its core, the system takes the full conversation history and the tutor’s response, then prompts an LLM, specifically, \texttt{GPT-4o} to assess whether the tutor has correctly identified a mistake in the student’s reasoning.

\begin{figure}
  \centering
  \includegraphics[width=1\linewidth]{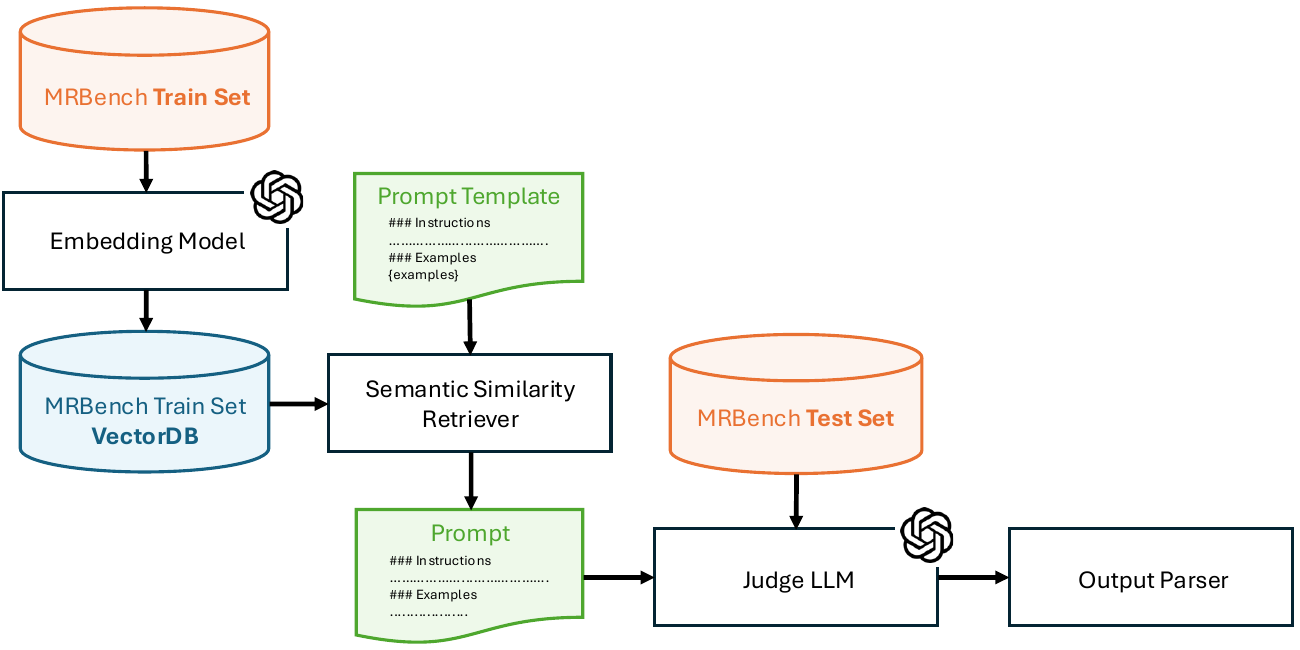}
    \caption{Pipeline of our final approach for mistake identification. The system takes tutor–student dialogue as input, retrieves relevant examples, constructs a structured prompt, and uses LLM to predict whether a mistake is identified. The output is parsed and saved.}
    \label{fig:arch}
\end{figure}

Figure~\ref{fig:arch} outlines the system architecture. We begin by embedding the conversation history and tutor responses from the \textsc{MRBench} training set using the \texttt{OpenAI Embedding Model}. These embeddings are stored in a persistent vector database using \texttt{ChromaDB}. With this setup, we construct a few-shot prompt template and use the LLM itself as a “judge” on the test data. At inference time, the system retrieves the top-$k$ semantically similar examples and integrates them into the prompt.

Each prompt includes detailed labeling instructions, definitions for all possible labels (\texttt{Yes}, \texttt{No}, \texttt{To some extent}), and the full dialogue context. (see Appendix \ref{app:prompt} for more information). To ensure clear and structured outputs, we use a \texttt{PydanticOutputParser} that enforces a strict schema and reliably extracts the label from the LLM’s response. The pipeline also supports retries and incremental saving, making it robust and efficient for large-scale processing.

By combining relevant examples with a powerful instruction-following model, this method allows for nuanced mistake identification beyond simple classification. It requires no fine-tuning, generalizes well to new inputs, and showed improvements in both accuracy and qualitative evaluations compared to baseline methods. This highlights the effectiveness of prompt-based, retrieval-augmented approaches in educational and feedback-driven NLP tasks.

\section{Dataset}

We use the dataset introduced by \citet{maurya-etal-2025-unifying}, which includes both development and test splits. The development set consists of \textit{300} dialogues from \citet{macina-etal-2023-mathdial} and \citet{wang-etal-2024-bridging}, each ending with a student utterance that reflects confusion or a mistake. Tutor responses, generated by seven large language model systems and human tutors (one in MathDial \citet{macina-etal-2023-mathdial}, expert and novice in Bridge), are annotated along four pedagogical dimensions: (1) Mistake Identification, (2) Mistake Location, (3) Providing Guidance, and (4) Actionability. In total, the development set includes over \textit{2,480} annotated responses.

The test set contains \textit{200} dialogues with the same structure, but tutor identities are anonymized (for example, Tutor\_1, Tutor\_2), and no annotations are provided. This allows for blind evaluation of system outputs under the shared task setting.

\subsection{Pre-processing}

For the baseline systems, we apply extensive preprocessing to both the conversation history and tutor response texts. This includes converting text to lowercase, removing punctuation, stripping emojis, and cleaning URLs, HTML tags, and contractions. We also remove stopwords using the NLTK stopword list. All texts are passed through a unified normalization pipeline to reduce noise and ensure consistency. The labels for Mistake Identification are mapped to numeric values as follows: \texttt{No} $\rightarrow$ 0, \texttt{Yes} $\rightarrow$ 1, and \texttt{To Some Extent} $\rightarrow$ 2.

For our final approach, we additionally preprocess both the development and test sets so that each dialogue is reformatted into evenly paired exchanges between tutor and student, preserving the integrity of the back-and-forth interaction. During this process, we addressed two key issues. First, some conversations included greetings or closing phrases (e.g., “Hi”, “Thank you”) that did not contribute to the reasoning process. These were removed to maintain focus on educational content. Second, a few dialogues contained erroneous segments where the tutor responded to its own utterance without student input. These cases were consistently found to follow a correctly structured exchange and were manually removed (see Appendix~\ref{app:preprocessing} for examples).

This pre-processing step ensured a clean and consistent input format, enabling reliable downstream processing and model evaluation.
\section{Evaluation and Results}

We evaluated all four approaches on the \textbf{Track 1: Mistake Identification} test set using two evaluation schemes: \textbf{Strict} and \textbf{Lenient}, each reporting both \textit{Macro F1} and \textit{Accuracy}. In the strict setting, only exact matches with the gold labels are considered correct. In contrast, the lenient setting provides partial credit by treating \texttt{To some extent} as aligning with \texttt{Yes}, reflecting the fuzzy nature of pedagogical judgments in borderline cases. Table~\ref{tab:results} summarizes the results. 

\setlength{\tabcolsep}{3pt}
\begin{table*}[]
\renewcommand{\arraystretch}{1.2}
\centering
    \begin{tabular}{L{7cm} R{2cm} R{2cm} R{2cm} R{2cm}}
    \toprule
     \textbf{Approach} & \textbf{Strict F1} & \textbf{Strict Acc} & \textbf{Lenient F1} & \textbf{Lenient Acc}\\
     \midrule
     Approach 1 (ML Ensemble) & 0.446 & 0.657 & 0.637 & 0.754 \\
     Approach 2 (Token-Level Attention) & 0.571 & 0.765 & 0.777 & 0.865 \\
     Approach 3 (CLS + MLP)   & 0.583 & 0.809 & 0.805 & 0.888 \\
    Approach 4 (Few-shot LLM + Retrieval) & \textbf{0.584} & \textbf{0.827} & \textbf{0.814} & \textbf{0.897} \\ \hline
    \end{tabular}
    \caption{Performance of all four approaches on the BEA 2025 Mistake Identification test set under strict and lenient evaluation settings.}
    \label{tab:results}
\end{table*}

As expected, the first baseline using pooled token embeddings and an ensemble of traditional classifiers (Approach 1) offered a modest starting point. This method, while straightforward, lacked the capacity to fully capture the nuances in dialogue-based reasoning.

Introducing token-level attention in Approach 2 led to a notable jump in performance. This suggests that modeling fine-grained interactions between the student’s dialogue and the tutor’s response helps the model better identify whether a mistake was correctly addressed. However, while this approach added depth to the representation, it still relied on relatively shallow modeling of the context.

Approach 3, which used frozen \texttt{[CLS]} embeddings from a sentence transformer combined with an MLP classifier, further improved performance. This indicates that sentence-level semantic representations, especially when paired with a focused classification head, can offer a stronger understanding of the overall pedagogical intent.

Our final method, Approach 4, which frames the task as a retrieval-augmented prompting problem with \texttt{GPT-4o}, achieved the best performance across all metrics. By retrieving semantically similar examples and using detailed, schema-guided prompts, the system benefited from both contextual grounding and the powerful instruction-following capabilities of modern LLMs. Notably, it showed strong results in both strict and lenient settings, highlighting its ability to make fine distinctions while still handling ambiguity in borderline cases effectively. Our final submission, achieved an official leaderboard rank of \textbf{37\textsuperscript{th}} among all participants.

\section{Conclusion}

We developed and evaluated four approaches for the BEA 2025 Shared Task \textbf{Track 1: Mistake Identification}, culminating in a retrieval-augmented few-shot prompting system using \textsc{GPT-4o}. While our initial baselines used traditional classifiers over pretrained embeddings, the final system reframed the task as a structured judgment problem, combining semantically retrieved examples, instruction-driven prompting, and schema-constrained output parsing.

This approach consistently outperformed all baselines in both strict and lenient evaluations, achieving a strict Macro F1 of \textit{0.584} and a lenient accuracy of \textit{0.897}. It was particularly effective at capturing nuanced pedagogical feedback, highlighting the strength of LLM-based reasoning when guided by relevant context. Our submission ranked 37\textsuperscript{th} on the official leaderboard, demonstrating the competitiveness of our method.

These results show that retrieval-augmented prompting offers a scalable and effective solution for assessing complex teaching behaviors in AI tutors. Future work could explore more adaptive example selection, multi-turn consistency, and alignment with broader goals such as helpfulness and instructional fairness.

\section*{Limitations}

While our final system achieved the best performance among all submitted approaches, it still has several limitations that suggest promising directions for future work.

\paragraph{Limited Diversity in Retrieved Examples}
The effectiveness of our retrieval-augmented prompting pipeline depends heavily on the quality and coverage of the example pool. Since we rely on a fixed set of annotated training examples, the system may struggle with out-of-distribution dialogues or question types that are underrepresented in the retrieval set. Moreover, retrieval is based solely on static embedding similarity from OpenAI embeddings, without adapting to the context or emphasizing specific pedagogical traits.

\paragraph{Lack of Multi-Turn Dialogue Modeling}
Each input is treated as a standalone conversation-response pair, with no memory of earlier tutor turns or evolving dialogue context. This limits the system’s ability to track learning progression or take prior feedback into account. Modeling dialogue history explicitly—through dialogue state tracking or memory-based retrieval—could improve consistency and pedagogical depth in multi-turn interactions.

\paragraph{Simplified Output Format}
Although the use of a structured parser ensures consistency, it restricts the model to selecting a single label per example. It does not capture uncertainty, nuanced justifications, or cases where multiple labels might apply. Extending the output to include rationales or confidence scores could make evaluations more informative and reflective of real-world ambiguity.

\paragraph{Scalability and Cost Constraints}
Inference with frontier models like GPT-4o is computationally intensive and dependent on external APIs, which introduces latency, cost, and rate-limit challenges. These constraints pose barriers to deployment in low-resource settings or real-time tutoring applications, where efficiency and scalability are critical.

\bibliography{custom}

\begin{thebibliography}{4}
\providecommand{\natexlab}[1]{#1}

\bibitem[{Kochmar et~al.(2025)Kochmar, Maurya, Petukhova, Srivatsa, Tack, and Vasselli}]{kochmar-etal-2025-bea}
Ekaterina Kochmar, Kaushal~Kumar Maurya, Kseniia Petukhova, KV~Aditya Srivatsa, Anaïs Tack, and Justin Vasselli. 2025.
\newblock Findings of the {BEA} 2025 shared task on pedagogical ability assessment of {AI}-powered tutors.
\newblock In \emph{Proceedings of the 20th Workshop on Innovative Use of NLP for Building Educational Applications}.

\bibitem[{Macina et~al.(2023)Macina, Daheim, Chowdhury, Sinha, Kapur, Gurevych, and Sachan}]{macina-etal-2023-mathdial}
Jakub Macina, Nico Daheim, Sankalan Chowdhury, Tanmay Sinha, Manu Kapur, Iryna Gurevych, and Mrinmaya Sachan. 2023.
\newblock \href {https://doi.org/10.18653/v1/2023.findings-emnlp.372} {{M}ath{D}ial: A dialogue tutoring dataset with rich pedagogical properties grounded in math reasoning problems}.
\newblock In \emph{Findings of the Association for Computational Linguistics: EMNLP 2023}, pages 5602--5621, Singapore. Association for Computational Linguistics.

\bibitem[{Maurya et~al.(2025)Maurya, Srivatsa, Petukhova, and Kochmar}]{maurya-etal-2025-unifying}
Kaushal~Kumar Maurya, Kv~Aditya Srivatsa, Kseniia Petukhova, and Ekaterina Kochmar. 2025.
\newblock \href {https://aclanthology.org/2025.naacl-long.57/} {Unifying {AI} tutor evaluation: An evaluation taxonomy for pedagogical ability assessment of {LLM}-powered {AI} tutors}.
\newblock pages 1234--1251, Albuquerque, New Mexico. Association for Computational Linguistics.

\bibitem[{Wang et~al.(2024)Wang, Zhang, Robinson, Loeb, and Demszky}]{wang-etal-2024-bridging}
Rose Wang, Qingyang Zhang, Carly Robinson, Susanna Loeb, and Dorottya Demszky. 2024.
\newblock \href {https://doi.org/10.18653/v1/2024.naacl-long.120} {Bridging the novice-expert gap via models of decision-making: A case study on remediating math mistakes}.
\newblock pages 2174--2199, Mexico City, Mexico. Association for Computational Linguistics.

\end{thebibliography}

\newpage

\appendix

\section{Preprocessing Examples}
\label{app:preprocessing}

This appendix list some issues which are fixed during pre-processing of dataset.

In the following example from the development set, the initial student message is a casual greeting that disrupts the expected alternating structure of the dialogue. To maintain structural integrity and ensure an even number of turns between tutor and student, such non-essential messages are removed during preprocessing.

\begin{minted}{python}
[
'Student: okey',
'Tutor: What is 25 minus 18?', 
'Student: 8'
]
\end{minted}

In the example below, the final tutor response erroneously mimics the student’s explanation, as if the tutor is responding to itself rather than engaging with the student. This type of error breaks the natural flow of the dialogue and was manually identified and removed during preprocessing to ensure accurate tutor-student interaction.

\begin{minted}{python}
[
'Tutor: Hi, could you please provide a step-by-step solution for the question below? The question is ...', 
'Student: Samantha buys 4 toys at $12.00 each. For each pair of toys...', 
'Tutor: I added the two amounts together to get a total of $36.00 + $6.00 = $42.00.'
]
\end{minted}

In cases like the example below, the tutor’s prompt is split across multiple turns, breaking the intended question into separate messages. To preserve the coherence of the dialogue and maintain a consistent turn-taking structure, such fragmented tutor responses are merged into a single utterance by concatenating the strings.

\begin{minted}{python}
[
'Tutor: Hi, could you please provide a step-by-step solution for the question below? The question is: Tyson decided to make muffaletta sandwiches for ..., 
'Tutor: How many pounds of meat are needed for each sandwich?', 
'Student: Each sandwich requires 1 pound of meat and 1 pound of cheese.', 
'Tutor: What is the cost of 1 pound of meat?', 'Student: The cost of 1 pound of meat is $7.00.'
]
\end{minted}

\newpage

\onecolumn
\section{Prompt Engineering}
\label{app:prompt}
\begin{figure*}[!h]
\begin{minted}{python}
"""
You will be shown a short educational "Conversation" between a tutor and a student, including the student's solution and the tutor's follow-up "Response". Your task is to judge whether the tutor's response successfully **identifies a mistake** in the student's reasoning.

### Instructions
1. Read the entire dialogue to understand the context of the student's solution.  
2. Focus on whether the tutor's response explicitly or implicitly calls out an error.  
3. Reply **only** with one of the labels: `Yes`, `To some extent`, or `No`.  

### Labels
- `Yes`: The mistake is clearly identified/recognized in the tutor's response. The tutor implicitly or explicitly points out the error in the student's reasoning.
- `No`: The tutor's response does not identify any mistake in the student's reasoning. The tutor's response is either irrelevant or does not address the student's solution.
- `To some extent`: The tutor's response suggests that there may be a mistake, but it sounds as if the tutor is not certain.

### Format Instructions:
{format_instructions}
Return only the classification label without any additional commentary or extraneous details.

### Examples
{examples}

## Mistake Identification
### Conversation
{conversation}

### Response
{response}
"""
\end{minted}
\centering
\caption{Prompt for LLM which is used a judge in Retrieval-Augmented Few-shot classification approach}
\label{fig:enter-label}
\end{figure*}

\end{document}